\pdfoutput=1

\documentclass[11pt]{article}

\usepackage[final]{acl}

\usepackage{times}
\usepackage{latexsym}

\usepackage[T1]{fontenc}

\usepackage[utf8]{inputenc}

\usepackage{microtype}

\usepackage{inconsolata}

\usepackage{graphicx}

\usepackage{hyperref}
\usepackage{url}

\usepackage{amsmath}
\usepackage{amsthm}
\usepackage{amsfonts}
\usepackage{bm}
\usepackage{thmtools,mathtools}
\usepackage{thm-restate}
\usepackage[T1]{fontenc}
\usepackage{booktabs}
\usepackage{multirow}
\usepackage{booktabs}
\usepackage{verbatim}
\usepackage{mdwlist}
\usepackage[ruled,norelsize,linesnumbered]{algorithm2e}
\usepackage{dsfont}
\usepackage[shortlabels]{enumitem}
\usepackage{caption}
\usepackage{subcaption}
\usepackage{array}
\usepackage{pifont}
\usepackage{CJKutf8}

\usepackage{multicol}
\usepackage{float}
\usepackage{lipsum}
\usepackage[font=small]{caption}
\usepackage{svg}
\usepackage{wrapfig}
\theoremstyle{definition}

\newcommand{\E}{\mathbb{E}}

\makeatletter
\newcommand{\removelatexerror}{\let\@latex@error\@gobble}
\makeatother

\newcommand{\codename}{\textit{fLSA}\xspace}

\title{fLSA: Learning Semantic Structures in Document Collections\\Using Foundation Models}

\author{Weijia Xu, Nebojsa Jojic \& Nicolas Le Roux \\
Microsoft Research\\
Redmond, WA 98052, USA \\
\texttt{\{weijiaxu,jojic,nleroux\}@microsoft.com} \\
}

\begin{document}

\maketitle

\begin{abstract}
\looseness=-1
Humans can learn to solve new tasks by inducing high-level strategies from example solutions to similar problems and then adapting these strategies to solve unseen problems. Can we use large language models to induce such high-level structure from example documents or solutions?
We introduce \codename, a foundation-model-based Latent Semantic Analysis method that iteratively clusters and tags document segments based on document-level contexts. These tags can be used to model the latent structure of given documents and for hierarchical sampling of new texts. Our experiments on story writing, math, and multi-step reasoning datasets demonstrate that \codename tags are more informative in reconstructing the original texts than existing tagging methods. Moreover, when used for hierarchical sampling, \codename tags help expand the output space in the right directions that lead to correct solutions more often than direct sampling and hierarchical sampling with existing tagging methods.
\end{abstract}

\section{Introduction}

\begin{figure*}[!ht]
\centering
\includegraphics[width=\textwidth]{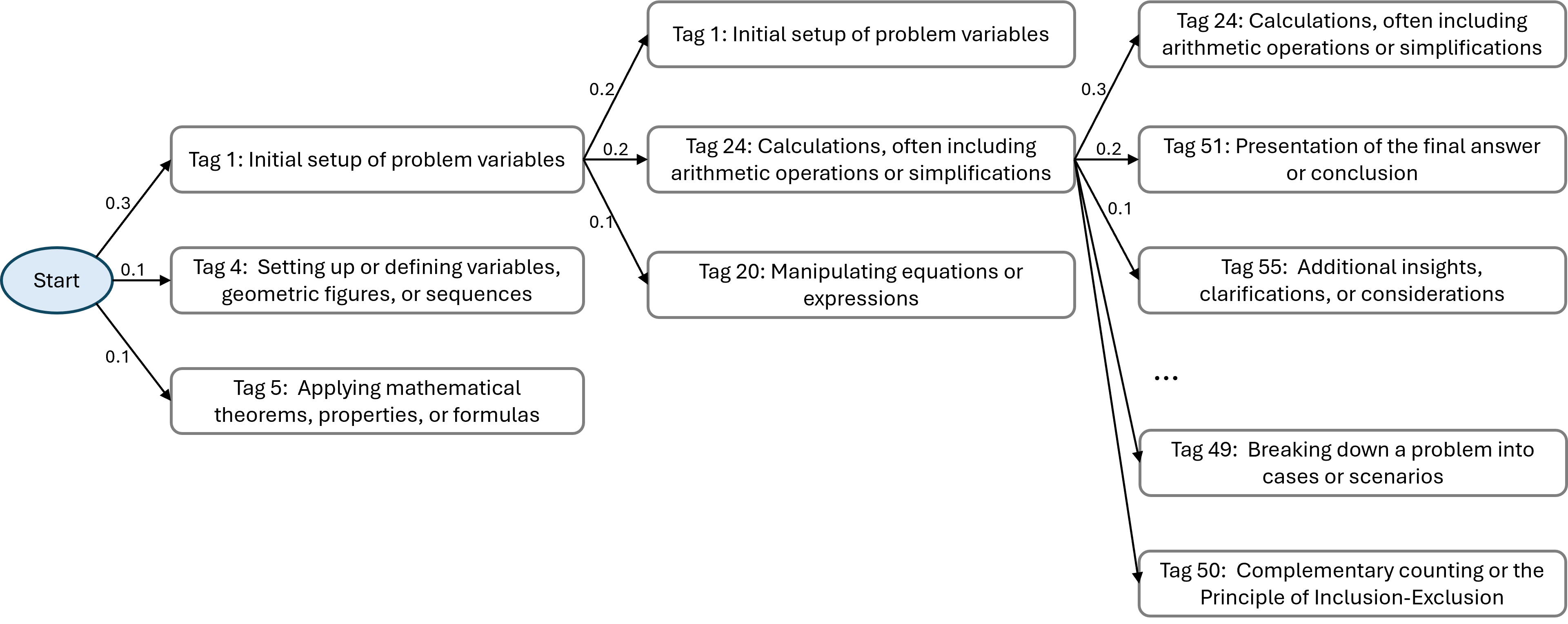}
\caption{Visualizing the bigram dynamic model over the latent tags learned on MATH solutions. For each tag, we list the three most probable next tags based on the transition probabilities~$p(t_k|t_{k-1})$. The transition probabilities are annotated on the arrows. For Tag~24, we also list~two example next tags outside the top-3 choices with transition probabilities~$p \approx 0.01$.}
\label{fig:dynamics_MATH}
\end{figure*}

\begin{figure}[!ht]
\centering
\includegraphics[width=0.5\textwidth]{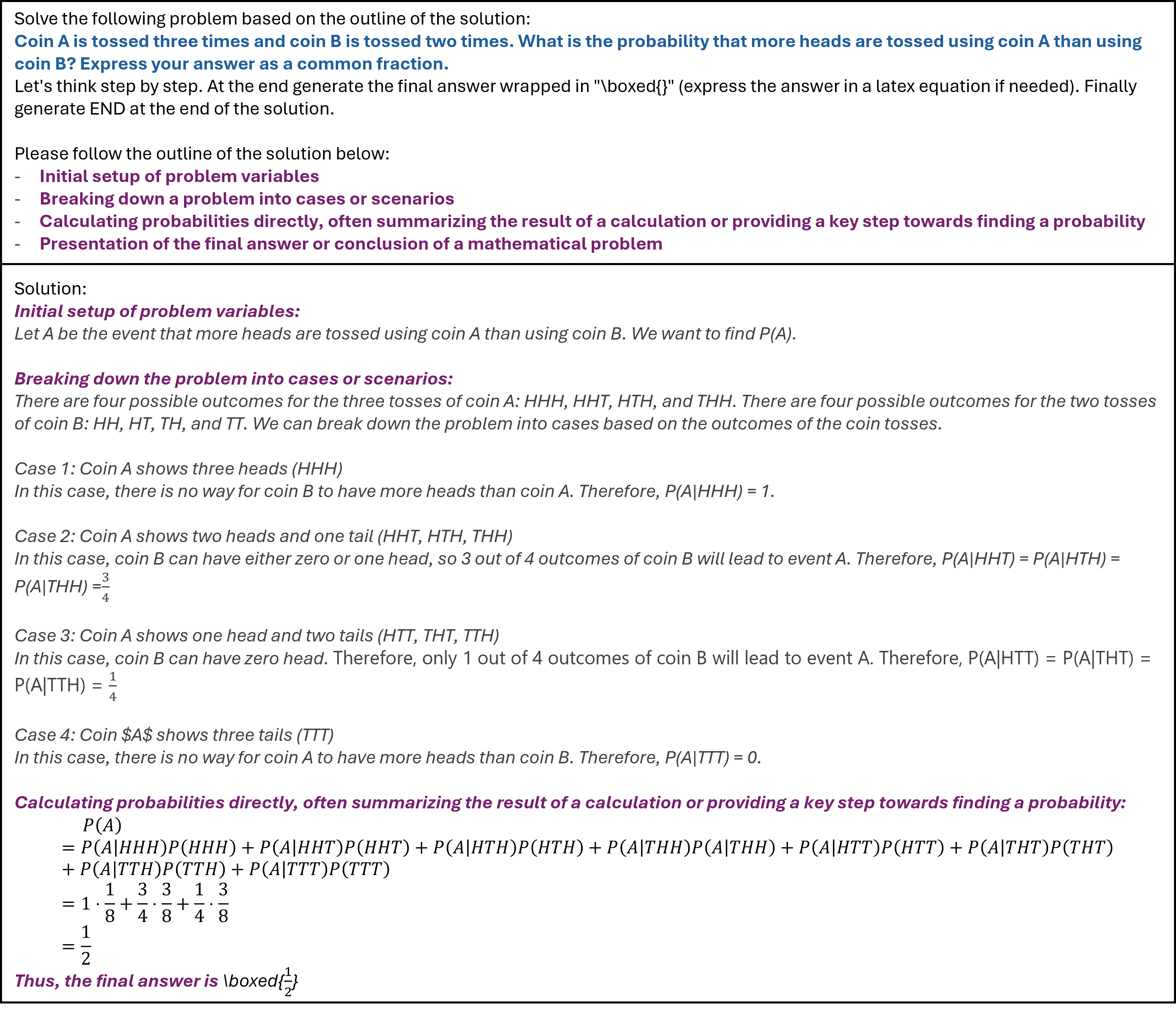}
\caption{An example of using the sampled tag sequence as an outline~(in purple) to aid an LLM in generating a solution~(italicized) to the given problem~(in blue).}
\label{fig:hiergen_example}
\end{figure}

\looseness=-1
Large language models~(LLMs) have shown impressive performance on a wide range of tasks, such as reasoning~\citep{suzgun2022challenging,liu2023evaluating}, math problem solving~\citep{wu2023empirical}, and open-ended text generation tasks~\citep{katz2024gpt,dubey2024llama,openai2024gpt4technicalreport}. Given natural language instructions or in-context examples with chain-of-thought steps, LLMs can adapt quickly to a new task and achieve outstanding performance on challenging tasks that require multi-step reasoning or planning~\citep{wei2022chain}. 
However, such methods typically rely on humans to induce the common strategy for solving a type of problems and demonstrate the strategy through few-shot chain-of-thought prompting. By contrast, humans learn to solve a new type of problems by analyzing some example problems and their solutions, inducing the common strategies~(i.e. \textit{latent semantic structure}) underlying these problem solutions, and testing them out on the new problems. %

Inducing the latent semantic structure in a set of documents can be modeled as an unsupervised clustering and tagging problem, where given a set of coarsely segmented documents, we cluster the text segments that share common characteristics into the same set and assign a tag to each set of segments. Based on these segment tags, we can then uncover the latent structure by learning a dynamic model over the latent tags and their transition probabilities in the document set. As an example, Figure~\ref{fig:dynamics_MATH} shows a dynamic model over learned tags in mathematical solutions. Such dynamic models can help humans better understand and analyze large collections of documents. They also encode more generalizable information compared to few-shot examples, providing a useful guide for LLMs to solve new problems without manual intervention (as shown by the example in Figure~\ref{fig:hiergen_example}). Additionally, they can also aid in searching algorithms on complex reasoning tasks~\citep{guan2025rstarmathsmallllmsmaster} through hierarchical sampling: one can sample from the dynamic model over latent tags as an outline for the actual solution steps to explore more diverse solution paths during the rollout stage.

In this paper, we introduce \codename, an iterative algorithm that alternatively clusters and tags document segments using LLMs based on segment- and document-level contexts. \codename combines the merits of traditional topic modeling approaches such as Latent Semantic Analysis~(LSA)~\citep{hofmann1999lsa} and LLM-based approaches, and captures shared semantic features among text segments more effectively.
We evaluate~1) the informativeness of \codename tags by measuring how well they help reconstruct the original text spans, and~2) their usefulness in expanding the search space in the right directions by measuring the Hits@K accuracy of the generated solutions through hierarchical sampling using the tags.  
Experiments on story writing, math and multi-step reasoning datasets show that \codename leads to higher reconstruction likelihood than existing tagging approaches. Furthermore, on math and reasoning tasks, hierarchical sampling using \codename tags helps expand the output space in the right directions more effectively than both direct sampling and existing tagging methods.

\section{Related Work}

\subsection{Document Segmentation and Labeling}
To model the structure and topic shifts in a document, prior work has introduced unsupervised document segmentation and labeling approaches that leverage term co-occurrence features~\citep{hearst1997text}, co-occurrence shifts in topic vectors~\citep{riedl-biemann-2012-topictiling}, lexical features and word embeddings~\citep{glavas-etal-2016-unsupervised}. These approaches focus mostly on lexical features which are limited in modeling the high-level semantic structure of documents. On the other hand, Neural-based approaches have the potential of modeling sentence-level semantics and document-level topic flows more effective, but rely heavily on supervised training samples in the target domain~\citep{koshorek-etal-2018-text,arnold-etal-2019-sector,zhang2019outline}. 
Our algorithm infers the structure of documents based on segment- and document-level contexts using LLMs in an unsupervised fashion.

\subsection{Topic Modeling}
\looseness=-1
Topic modeling is a widely used technique in natural language processing for uncovering hidden thematic structures in large text corpora. The most foundational methods in this domain include Latent Dirichlet Allocation~(LDA)~\citep{blei2003latent} and Latent Semantic Analysis~(LSA)~\citep{hofmann1999lsa,hofmann1999lsi,hofmann2001unsupervised}. Both methods represent each document as a bag of words and models word-document relationships using a mixture of latent topics, where each topic is represented by a list of top words. These algorithms are mathematically grounded, but typically rely on manual topic interpretation, which often leads to incorrect or incomplete labels~\citep{Gillings2022}. 
More recent work introduces neural topic models~\citep{NVDM,ETM,srivastava2017autoencodingvariationalinferencetopic}, which combine traditional topic models with word embeddings. These models have shown improved performance in handling large and complex vocabularies. However, they still model each document as a bag of words, disregarding the sentence- and document-level semantics. Additionally, the resulting topics are represented either by semantic vectors or lists of closest words, which still rely on manual interpretation. 
Furthermore, studies have shown that incorporating expert knowledge in topic modeling improves over traditional unsupervised methods~\citep{LEE201728}.

Moreover, the advent of large language models~(LLMs) has led to LLM-based topic modeling approaches. \citet{LiZZ2023} propose to use LLMs for topic labeling based their top terms produced by traditional topic models. For short text spans, however, the bag-of-words representation of texts provides limited information for topic modeling. \citet{akash2023letpretrainedlanguagemodels} address the issue by extending each text span into longer sequences using LLMs and extracting topics from the extended texts using neural topic models. Futhermore, \citet{pham-etal-2024-topicgpt,Wang2023PromptTopic,mu-etal-2024-large} propose prompt-based techniques to generate, merge, and assign topics using LLMs. 
These approaches leverage the domain knowledge embedded in LLMs and produce more interpretable topics based on sentence or document-level contexts beyond bag of words.

However, the generate-and-merge approach limits the model's potential for discovering shared features among various text spans across documents of different themes and often leads to overly abstract, thematical topics, especially on a large-scale document collection. We propose \codename, which combines the merits of traditional LSA, which uses an iterative EM algorithm to model topic and text distributions, and LLM-based approaches.

\section{Approach}

\begin{figure*}[!ht]
\centering
\includegraphics[width=\textwidth]{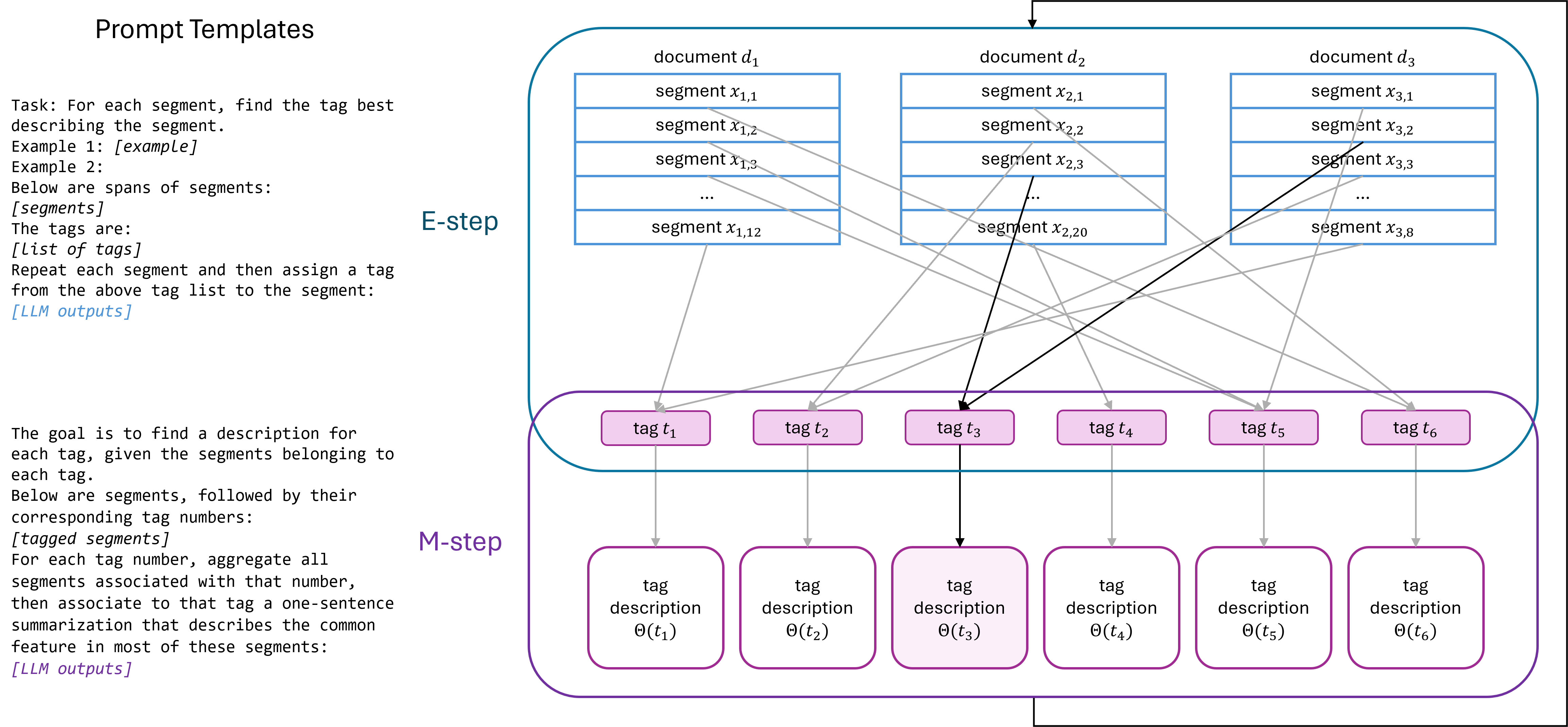}
\caption{An illustration of the E-step and M-step in \codename. At the E-step, we assign each text segment to a tag through prompting given the tag descriptions at the previous iteration. At the M-step, we prompt the LLM to generate new tag descriptions based on the segments assigned to each tag at the E-step.}
\label{fig:em_algorithm}
\end{figure*}

We propose \codename, a foundation-model-based EM algorithm that learns the latent tags on a set of segmented documents. We draw inspiration from the traditional Probabilistic Latent Semantic Analysis and use iterative EM steps to learn the latent tags that maximize the estimated likelihood of segmented documents.

\subsection{Probabilistic Latent Semantic Analysis~(PLSA)}
PLSA models the distribution over words~$w$ in a document~$d$ as a mixture of conditionally independent multinomial distributions, each such distribution representing a \emph{topic}~$t$. This generative model of words in a document is usually expressed mathematically in terms of the distribution:
\begin{equation}
    p_\Theta(w | d) = \sum_t p_\Theta(t | d) p_\Theta(w | t),
\end{equation}
which can be sampled by first sampling a topic~$t$ for the given document~$d$ from~$p_\Theta(t | d)$ and then sampling words conditioned on the topic from~$p_\Theta(w | t)$.  $\Theta$ represents the parameters of the PLSA model. PLSA aims to find~$\Theta$ that maximizes the log-likelihood of words in all documents:
\begin{equation}
    \mathcal{L} = \sum_{d, w} \log \sum_t p_\Theta(t | d) p_\Theta(w | t)
\label{eq:plsa_objective}
\end{equation}

To estimate the parametric distributions~$p_\Theta(t | d)$ and~$p_\Theta(w | t)$, PLSA relies on an EM algorithm, which is an iterative method to find the maximum likelihood estimate of parameters in statistical models. Specifically, an EM iteration alternates between an expectation~(E) step and a maximization (M) step. At iteration~$i$, the E-step estimates the posterior distribution~$p_{\Theta_{i-1}}(t | w, d)$ of topics~$t$ conditioned on each document~$d$ and word~$w$ in it based on fixed parameters~$\Theta_{i-1}$ from the previous iteration:
\begin{equation}
    p_{\Theta_{i-1}}(t | w, d) = \frac{p_{\Theta_{i-1}}(t | d) p_{\Theta_{i-1}}(w | t)}{\sum_{t'} p_{\Theta_{i-1}}(t' | d) p_{\Theta_{i-1}}(w | t')}
\end{equation}
The M-step optimizes the parameters~$\Theta$ such that the expectation of the log-likelihood~$p_\Theta(w | d)$ of words in each document given~$t$ sampled from the estimated posterior~$p_{\Theta_{i-1}}(t | w, d)$ is maximized:
\begin{equation}
    \arg\max_\Theta \sum_{d, w} \E_{t \sim  p_{\Theta_{i-1}}(t | w, d)} \log p_\Theta(t | d) p_\Theta(w | t)
\end{equation}
Theoretically, each EM iteration will yield a larger likelihood in Eq~\ref{eq:plsa_objective} until it converges to a local maximum. In topic modeling literature, various generalized EM variants exist, including the ones that approximate the posterior distribution with a small number of samples, or just the mode of it, and which alter the parameters so that they do not necessarily maximize the likelihood under the posterior, but simply improve it.

\subsection{Foundation-Model-Based LSA~(fLSA)}
We introduce \codename, which learns the latent \emph{tags}~(similar to \emph{topics} in LSA)\footnote{We use the terminology \emph{tag} instead of \emph{topic} in our algorithm because they may cover shared characteristics among document segments beyond topics (see the example tags in Figure~\ref{fig:dynamics_MATH}).} on a set of segmented documents~$d = (x_1, x_2, ..., x_L)$, where the document~$d$ is segmented into~$L$ segments~$x_k$. 
A core difference between \codename and PLSA is that PLSA models the generative probability of each word in a document independently, while \codename models the probability of the sequence of words~$(w_1, w_2, ..., w_n)$ in each text segment~$x_k$ jointly as~$p_\Theta(w_1, w_2, ..., w_n | t)$. Moreover, PLSA models the distribution over tags~$p_\Theta(t | d)$ for each document independently of other documents, while \codename models the distribution over tags~$t$ conditioned not only on current segment~$x_k$ but also on the document~$d$. 

To express the difference mathematically, in \codename, the generative model of a segment~$x_k=w_{1..n}$ in a document~$d$ can be written as:
\begin{equation}
    p_\Theta(w_{1..n}| x_k, d) = \sum_t p_\Theta(t | x_k, d) p_\Theta(w_{1..n} | t),
\label{eq:fplsa_objective}
\end{equation}
which can be sampled by first sampling a tag~$t$ for the current segment~$x_k$ in document~$d$ and then sampling the word sequence~$w_{1..n}$ for that segment given the tag.

Another core difference between \codename and PLSA is that we model the parametric distributions~$ p_\Theta(t | x_k, d)$ and~$p_\Theta(w_{1..n} | t)$ using an LLM with frozen parameters, and the tunable ``parameters''~$\Theta$ in \codename are the \emph{textual} description~$\Theta(t)$ for each tag~$t$ and the tag assignment for each segment.

Analogously to the (generalized) EM algorithms for traditional topic models, we are seeking ~$\Theta$ that corresponds to high likelihood of the word sequence in each document:
\begin{equation}
    \mathcal{L} = \sum_{d, x_k} \log \sum_t p_\Theta(t | x_k, d) p_\Theta(w_{1..n} | t)
\end{equation}

Our iterative EM steps are shown in Figure~\ref{fig:em_algorithm}.
At the E-step in iteration~$i$, we approximate the posterior distribution~$p_{\Theta_{i-1}}(t | w_{1..n}, x_k, d)$ of tags~$t$ for each segment~$x_k=w_{1..n}$ in document~$d$ by prompting the LLM to greedily assign a tag given the tag descriptions~$\Theta_{i-1}(t)$ from the previous iteration, the current segment~$x_k=w_{1..n}$ and neighbouring segments~$(x_{k-W/2}, x_{k+1-W/2}, ..., x_{k-1+W/2}, x_{k+W/2})$ as document-level context, where~$W$ is the context window size.\footnote{At the first iteration, since the tag descriptions are empty, we assign tags randomly.}
At the M-step, in lieu of maximizing (or just improving) the expected log-likelihood~$p_\Theta(w_{1..n} | x_k, d)$ of words in each segment given the tag assignments from the E-step,
\begin{equation}
\begin{split}
    \arg\max_\Theta \sum_{d, x_k} &\E_{t \sim  p_{\Theta_{i-1}}(t | w_{1..n}, x_k, d)} \\
    &\log p_\Theta(t | x_k, d) p_\Theta(w_{1..n} | t),
\end{split}
\end{equation}
we obtain updated tag descriptions~$\Theta(t)$ by inviting the LLM itself to summarize the segments assigned to the tag $t$: We aggregate the segments assigned to tag~$t$ and prompt the LLM to generate a tag description that best summarizes what these segments share in common (Fig. \ref{fig:em_algorithm}).

\section{Experimental Setup}

\subsection{Datasets}
\looseness=-1
We evaluate \codename against various baselines on WritingPrompts~(a story writing dataset~\citep{fan-etal-2018-hierarchical}), MATH~(which contains math problems and the corresponding solution texts~\citep{hendrycks2021math}), and Big-Bench Hard~(BBH) benchmark~(which contains diverse types of reasoning problems and their solutions~\citep{suzgun2022challenging}). We set the number of tags to~100 for WritingPrompts and MATH, and~50 for BBH~(see the Appendix for more details).

\subsection{Evaluation Metrics}

\paragraph{Reconstruction Likelihood}
To measure the informativeness of learned tags (either through \codename or a baseline algorithm), we measure the reconstruction log-likelihood of the test documents (stories in the test set of WritingPrompts or problem solutions in the test set of MATH) conditioned on the tags.

Specifically, for each test case~$x_k$, which is a segment randomly sampled from a test document~$x_{1...L}$~(randomly sampled from the test corpus), we approximate the reconstruction log-likelihood of~$x_k$ given latent tags~$t_k$ predicted given~$x_k$ and its neighboring segments under the LLM:
\begin{equation}
    \E_{t_k \sim p_{LLM}(t | x_k, d)}[\log p_{LLM}(x_k | x_{1...k-1}, t_k)]
\end{equation}

Specifically, we first sample~$S$ alternative segments at position~$k$ independently by~$\{ \tilde{x}_k^{(1)}, \tilde{x}_k^{(2)}, ..., \tilde{x}_k^{(S)} \} \sim p_{LLM}(\cdot | x_{1...k-1})$.
Next, we conduct~$T$ repeated experiments to approximate the log-likelihood of~$x_k$ given the previous segments~$x_{1…k-1}$ and the tag~$t_k$ predicted on~$x_k$ under the LLM. Each time, we randomly sample~$C$ alternative segments from~$\{ \tilde{x}_k^{(1)}, \tilde{x}_k^{(2)}, ..., \tilde{x}_k^{(S)} \}$ and put it together with~$x_k$ (in randomly shuffled order) as options and ask the LLM which one is the true continuation conditioned on~$x_{1...k-1}$ and~$t_k$. 
Based on the number of times~(denoted as $c_k$) that the LLM chooses~$x_k$ as the true continuation among all~$T$ experiments, we estimate the reconstruction log-likelihood with alpha-smoothing~($\alpha=0.1$):
\begin{equation}
\begin{split}
    &\E_{t_k \sim p_{LLM}(t | x_k, d)}[\log p_{LLM}(x_k | x_{1...k-1}, t_k)]\\
    &= \log\frac{c_k + \alpha}{T + \alpha S}
\end{split}
\end{equation}
As a baseline, we compare the reconstruction log-likelihood with the log-likelihood computed the same way as above but without conditioning on any tags:
\begin{equation}
    \E[\log p_{LLM}(x_k | x_{1...k-1})] = \log\frac{c_k' + \alpha}{T + \alpha S}
\end{equation}
where~$c_k'$ is the number of times that the LLM chooses~$x_k$ as the true continuation among~$T$ experiments, which is computed the same way as above except that when asking the LLM to choose the true continuation, we only provide the previous text segments~$x_{1...k-1}$ without any tags. %

In our experiments, we evaluate the reconstruction log-likelihood of all methods on the same set of~1K randomly sampled test cases.

\paragraph{Hits@K Accuracy} 
\looseness=-1
To demonstrate that the learned tags can also help expand the search space in the right directions when searching for effective solutions to a complex reasoning task, we learn a dynamic model over the latent tags~(as shown by the example in Figure~\ref{fig:dynamics_MATH}) and use it for hierarchical sampling, where we first sample a sequence of tags as an outline and then sample the actual text based on the outline.
And then, we evaluate the Hits@K accuracy of hierarchical sampling with latent tags, and compare it with the Hits@K accuracy of direct sampling without tags. Specifically, for each problem, we sample~$K=50$ solutions independently from an LLM given the problem description either directly or through hierarchical sampling with latent tags. If any of the~$K$ solutions leads to the correct answer, it gets a score of 1, otherwise 0. Finally, we compute the average score over all testing problems.

For hierarchical sampling, we first sample a sequence of tags~$(t_1, t_2, ..., t_l)$ (up till the special tag <END>) with maximum length L using a bigram model learned on the training data (without conditioning on the test problem):
\begin{equation}
\begin{split}
    &p(t_1, t_2, ..., t_l) \\
    =& p(t_1) p(t_2 | t_1) ... p(t_l | t_{l - 1}) p(\text{<END>} | t_l)
\end{split}
\end{equation}
And then, we prompt the LLM to generate a solution to the given problem based on the tag sequence~$(t_1, t_2, ..., t_l)$ using the prompt template shown in Figure~\ref{fig:hiergen_example}.

\subsection{\codename Setup}
For the EM procedure, we set the maximum number of iterations to~30.\footnote{We found in our preliminary experiment that the learned tag descriptions become stable (with very little semantic changes) in less than 30 iterations.} At the E-step~(where the LLM assigns a tag to each segment conditioned not only on the current segment but also on neighbouring segments within the context window), we use a context window size of~2 on WritingPrompts and use unlimited context window~(such that the whole solution is used as context) on MATH and BBH. At the M-step, we randomly sample~10 segments assigned to each tag to update the tag description.

\subsection{Baselines}
\paragraph{TradLDA}
\looseness=-1
We compare our approach with the traditional Latent Dirichlet Allocation~(\textit{TradLDA}), a type of LSA algorithm designed to discover latent topics in a collection of text spans~\citep{blei2003latent}.

\paragraph{TradLDA+LLM}
As \citet{LiZZ2023} showed that the topic labels generated by LLMs based on the key terms learned through TradLDA are preferred more often than the original labels, we also include \textit{TradLDA+LLM} as a baseline. Specifically, we first learn the topics and the key terms for each topic using TradLDA, and then use GPT-4 to generate a description for each topic based on the key terms.

\paragraph{Prompting}
Recent work showed that, with appropriate prompts, LLMs are capable of directly generating topic labels given a set of text documents and condensing
overarching topics~\citep{pham-etal-2024-topicgpt,Wang2023PromptTopic,mu-etal-2024-large}. As a baseline, we adapt the approach (along with the prompts) in \citet{mu-etal-2024-large} to generate topic descriptions for each text segment.

\paragraph{GenOutline}
For Hits@K accuracy, we also include a two-step sampling baseline, where we first prompt the LLM to generate a multi-step outline for solving this type of problem and then prompt the LLM to generate the actual solution based on the problem description and the outline.

\subsection{Large Language Model Setup}
For clustering and tagging, we use GPT-4~\citep{openai2024gpt4technicalreport} and Qwen-2.5-7B~(a much smaller LLM introduced in~\citet{qwen2025qwen25technicalreport}). We also use GPT-4 to estimate the reconstruction log-likelihood. 
To measure Hits@K Accuracy, we use ChatGPT~(gpt-3.5-turbo; \citet{openai2023gpt4}) instead of GPT-4, because GPT-4 has achieved high accuracy on MATH and BBH~(e.g. 84\% on MATH~\citep{zhou2023solvingchallengingmathword}), possibly due to data contamination issues~\citep{deng-etal-2024-investigating,bubeck2023sparksartificialgeneralintelligence}. Thus, we use ChatGPT for solution sampling to show the potential of using learned tags to diversify the sampled outputs and improve the chance of finding a correct answer when the model cannot find it through direct sampling.\footnote{More details in the Appendix.}

\section{Results}
\label{sec:results}

\subsection{Reconstruction Likelihood}
\begin{table*}
    \centering
    \scalebox{1}{
    \begin{tabular}{l|ccccc}
    & No Tag & TradLDA & TradLDA+LLM & Prompting & \codename \\
    \midrule
    WritingPrompts & -4.81 & -3.75 & -4.12 & -3.62 & \textbf{-3.43} \\
    MATH-Num & -3.32 & -2.96 & -3.28 & -3.06 & \textbf{-2.64} \\
    MATH-All & -3.67 & -3.16 & -3.57 & -3.44 & \textbf{-2.94} \\\midrule
    \end{tabular}
    }
\caption{Reconstruction log-likelihood of \codename versus the baseline without tags~(\textit{No Tag}), traditional LDA~(\textit{TradLDA}), traditional LDA with LLM-generated tag descriptions~(\textit{TradLDA+LLM})~\citep{LiZZ2023}, and the prompting baseline~(\textit{Prompting})~\citep{mu-etal-2024-large} on \textit{WritingPrompts} story dataset, Number Theory dataset from MATH~(\textit{MATH-Num}), and the MATH~(\textit{MATH-All}) dataset.}
\label{tab:recon_loglikelihood}
\end{table*}

\begin{table*}[t!]
\centering
\scalebox{1}{
\begin{tabular}{m{0.45\textwidth}|m{0.45\textwidth}}
\toprule
\textbf{Key Terms} & \textbf{Tag Description}
\\\midrule
{nothing, get, life, else, light, across, best, ca, single, come, got, death, together, running, power, system, entire, could, control, everything} & {The words you've provided span a broad range of concepts, but they share a common denominator in that they can all be associated with themes commonly found in science fiction literature and media.} \\\midrule
{continued, surface, wait, raised, floor, slowly, give, new, sure, needed, around, also, face, body, fact, made, bitch, girl, guy, much} & {The words listed seem to be common English words that could appear in a wide range of contexts. However, given their generic nature, they could be particularly prevalent in narrative or descriptive writing, such as in fiction, storytelling, or personal narratives.} \\\midrule
\end{tabular}
}
\caption{Examples of key terms learned on short story segments in WritingPrompts through \textit{TradLDA} and the corresponding tag descriptions generated by GPT-4. Given only the key terms without context, the tag descriptions produced by GPT-4 are too generic to recover the original text spans.}
\label{tab:TradLDA+LLM_examples}
\end{table*}

\begin{table*}[t!]
\centering
\scalebox{1}{
\begin{tabular}{m{0.45\textwidth}|m{0.45\textwidth}}
\toprule
\textbf{Prompting Tags} & \textbf{\codename Tags}
\\\midrule
{Tag 1: Stories involving themes of sacrifice, duty, friendship, companionship, hope, and resilience in the face of crisis.} & {Tag 1: Scenes involving intense, often dangerous situations, like explosions, retreats, long nights, empty streets, fires, and storms.} \\\midrule
{Tag 2: Stories involving time travel, genetic irregularities, and strange creatures that feed on negative emotions.} & {Tag 2: The protagonist experiences surreal and unexpected events, often involving time travel or strange bodily functions, and narrates them in a casual, humorous tone.} \\\midrule
{Tag 3: Stories involving emotional moments and first hugs.} & {Tag 3: This tag is associated with story segments that feature intense emotional moments, often involving fear, anger, or distress, and frequently serve as turning points or climactic scenes in the narrative.} \\\midrule
\end{tabular}
}
\caption{Example tags learned on short story segments in WritingPrompts through Prompting versus \codename. Prompting tags are either too mixed~(e.g. Tag 1 and 2) or too generic~(e.g. Tag 3), while \textit{\codename} groups segments of similar themes into the same cluster and describes each cluster with detailed explanations and example plots.}
\label{tab:prompting_vs_ours_wp_examples}
\end{table*}

First, we compare the reconstruction log-likelihood of \codename with the \textit{No Tag} baseline (without conditioning on any tags). As shown in Table~\ref{tab:recon_loglikelihood}, conditioning on \codename tags helps predict the original texts: \codename brings~0.7--1.4 higher log-likelihood than the \textit{No Tag} baseline.

\textit{TradLDA} also brings higher reconstruction log-likelihood over the \textit{No Tag} baseline. However, since \textit{TradLDA} only captures word or term co-occurrences, it still underperforms \codename consistently on all three datasets. Moreover, \textit{TradLDA+LLM} fails to improve over \textit{TradLDA}. As shown by the examples in Table~\ref{tab:TradLDA+LLM_examples}, it is extremely challenging for LLMs and even humans to extract meaningful semantic information from the key terms learned on short text segments through \textit{TradLDA}, and the resulting tag descriptions are overly generic, making it challenging to reconstruct the original text segments accurately.

\looseness=-1
Compared with the Prompting baseline, \codename achieves~0.2--0.5 higher log-likelihood on all three datasets. We further compared the tags learned using Prompting versus \codename. As shown by the examples in Table~\ref{tab:prompting_vs_ours_wp_examples}, Prompting tends to merge unrelated topics into a mixed topic~(e.g. Tag 1 and 2), and the resulting topics become overly broad. Even for tags sharing a common theme, the descriptions often lack specificity and detail~(e.g. Tag 3). By contrast, \codename identifies segments with similar themes, groups them into a single cluster and produces more detailed tag descriptions with example plots.%

\subsection{Hits@K Accuracy}

\begin{table*}[!ht]
    \centering
\scalebox{1}{
    \begin{tabular}{l|cccccc}
    & No Tag & GenOutline & TradLDA & TradLDA+LLM & Prompting & \codename \\\midrule
    \multicolumn{7}{l}{\bf MATH} \\\midrule
    Algebra & 88.6 & 90.1 & {\bf 93.6} & 89.6 & 91.1 & 90.1 \\
    Counting & 61.3 & 60.4 & 69.8 & 65.1 & 69.8 & {\bf 70.8} \\
    Geometry & 53.1 & 55.2 & 58.3 & 57.3 & {\bf 62.5} & 60.4 \\
    InterAlgebra & 55.7 & 51.7 & 58.7 & 59.2 & {\bf 61.2} & {\bf 61.2} \\
    Number & 65.4 & 76.0 & 77.9 & 74.0 & 78.8 & {\bf 83.7} \\
    PreAlgebra & 74.2 & 79.1 & 81.3 & 81.3 & 84.6 & {\bf 89.0} \\
    PreCalculus & 42.2 & 46.8 & 51.4 & 46.8 & 49.5 & {\bf 55.0} \\\midrule
    {\bf Average} & 62.9 & 65.6 & 70.1 & 67.6 & 71.1 & {\bf 72.9} \\\midrule
    \multicolumn{7}{l}{\bf BBH} \\\midrule
    Date & 92.8 & 94.4 & 95.6 & 95.2 & 95.2 & {\bf 98.8} \\
    Formal & 45.2 & 61.2 & 65.6 & 52.8 & 57.2 & {\bf 93.2} \\
    Geometric & 70.8 & 76.8 & 83.6 & 84.0 & 80.0 & {\bf 87.6} \\
    Logical & 89.2 & 95.6 & 95.6 & 96.0 & 96.5 & {\bf 99.5} \\
    Movie & 84.8 & 88.0 & 92.8 & 92.0 & 93.2 & {\bf 95.2} \\
    ObjCount & 93.2 & 96.8 & 99.2 & {\bf 100.0} & {\bf 100.0} & 95.2 \\
    Penguins & 93.8 & 99.3 & 99.3 & {\bf 100.0} & 99.3 & 99.3 \\
    ReasonColored & 92.8 & 97.6 & 98.4 & 98.8 & 98.8 & {\bf 100.0} \\
    RuinNames & 64.8 & 74.8 & 69.6 & 70.0 & 80.0 & {\bf 93.6} \\
    TranslationError & 52.4 & 68.4 & 60.4 & 60.0 & 63.6 & {\bf 75.2} \\
    Temporal & 86.4 & 98.4 & 93.2 & 96.8 & 98.0 & {\bf 100.0} \\
    WordSort & 27.2 & 36.4 & 16.0 & 14.8 & 42.0 & {\bf 56.0} \\\midrule
    {\bf Average} & 74.5 & 82.3 & 80.8 & 80.0 & 83.7 & {\bf 91.1} \\\midrule
    \end{tabular}
}
\caption{Hits@K accuracy of \codename versus directly sampling without tags~(\textit{No Tag}), two-step sampling with LLM-generated outline~(\textit{GenOutline}), traditional LDA~(\textit{TradLDA}), traditional LDA with LLM-generated tag descriptions~(\textit{TradLDA+LLM})~\citep{LiZZ2023}, and the prompting baseline~(\textit{Prompting})~\citep{mu-etal-2024-large} on~12 challenging tasks from BBH benchmark~\citep{suzgun2022challenging} and~7 tasks from MATH~\citep{hendrycks2021math}.}
\label{tab:hitsk_acc}
\end{table*}

\looseness=-1
We further evaluate how the tags and semantic structure learned through \codename help expand the output space in the right directions that lead to correct solutions by measuring the Hits@K Accuracy of various sampling methods with or without tags. First, compared with direct sampling without using any tags, hierarchical sampling with \codename tags leads to significantly higher Hits@K accuracy by~+10.0 points on MATH and~+16.6 points on BBH on average. Additionally, we compare \codename with GenOutline, a two-step sampling approach where we prompt the LLM to generate an outline before generating the actual solution. GenOutline improves over direct sampling on most tasks, but still underperforms hierarchical sampling with \codename by~7--9 points. These results indicate that hierarchical sampling using tags derived from the domain-specific documents via \codename produces more effective output solutions, thereby increasing the likelihood of hitting the correct answer with K samples.

Next, we compare \codename with hierarchical sampling with existing tagging approaches. \codename tags expand the output space in the directions that lead to correct answers more often than TradLDA on~16 out of~19 tasks. It brings a significant improvement of~3--10 points over TradLDA.\footnote{We test significance using paired student's t-test with significance level~$\alpha=0.05$.} Similarly, compared with TradLDA+LLM, \codename achieves higher Hits@K Accuracy on~17 out of~19 tasks and improves the average accuracy by~5--11 points across BBH and MATH. 
Compared with the Prompting baseline, \codename achieves higher Hits@K Accuracy on~14 out of~19 tasks. Overall, hierarchical sampling with \codename tags improves Hits@K Accuracy significantly over existing tagging approaches by~2--11 points on average.

\subsection{Learning Tags with Smaller LLMs}
In addition to GPT-4, we also evaluate \codename using a smaller LLM \---\ Qwen-2.5-7B. We run \codename using Qwen-2.5-7B as the base model~(while the other hyper-parameters remain unchanged) and measure the Hits@K Accuracy of hierarchical sampling using the learned tags on BBH. We discover that the average accuracy drops by~5 points compared to the tags learned using GPT-4, but it still outperforms TradLDA and Prompting (using the much larger GPT-4 model) by~3--6 points. 

\subsection{Ablation Study}
We further examine how the number of tags learned through \codename influences its ability to expand the output space. Specifically, we compare the Hits@K Accuracy of hierarchical sampling with~20, 50, and~100 \codename tags on BBH tasks. Results show that the accuracy drops by~3 points when using~20 instead of~50 tags, whereas increasing the number of tags from~50 to~100 yields minimal change~(see the Appendix for detailed results). This suggests that learning a sufficient \---\ even redundant \---\ number of tags can be beneficial for effectively expanding the output space.

\section{Conclusion}
\looseness=-1
We introduced \codename, a foundation-model-based Latent Semantic Analysis method that aims to uncover the latent semantic structures in document collections by iteratively clustering and tagging document segments based on document-level contexts. 
Our experiments on story writing, math and multi-step reasoning tasks show that \codename tags are more informative in reconstructing the original texts than tags generated by existing tagging methods. \codename tags are also useful in expanding the output space via hierarchical sampling to increase the likelihood of discovering correct solutions to complex reasoning problems.
These results suggest the potential of \codename for generating effective task guidelines given some worked-out examples, along with hierarchical sampling and searching for problem solutions on challenging reasoning tasks.

\section{Limitations}
One limitation of \codename is that some of the tags produced by \codename may be semantically similar to each other, which can be ideally merged into a single tag.  This limitation could be addressed by incorporating a tag fusion step in the EM algorithm, which we leave for future work. In addition, although the \codename algorithm is agnostic to the LLM being used, we only test it on GPT-4 (which is one of the most powerful and widely used LLMs). Testing the algorithm on smaller models can be an interesting future work.

This work also has potential risks. One major risk is that the tags learned using \codename may reflect the undesirable biases within the LLM being used. Integrating bias detection and mitigation techniques within the algorithm could be useful for addressing the issue.

\bibliography{iclr2025_conference}

\begin{thebibliography}{36}
\providecommand{\natexlab}[1]{#1}

\bibitem[{Akash et~al.(2023)Akash, Huang, and Chang}]{akash2023letpretrainedlanguagemodels}
Pritom~Saha Akash, Jie Huang, and Kevin Chen-Chuan Chang. 2023.
\newblock \href {https://arxiv.org/abs/2310.15420} {Let the pretrained language models "imagine" for short texts topic modeling}.
\newblock \emph{Preprint}, arXiv:2310.15420.

\bibitem[{Arnold et~al.(2019)Arnold, Schneider, Cudr{\'e}-Mauroux, Gers, and L{\"o}ser}]{arnold-etal-2019-sector}
Sebastian Arnold, Rudolf Schneider, Philippe Cudr{\'e}-Mauroux, Felix~A. Gers, and Alexander L{\"o}ser. 2019.
\newblock \href {https://doi.org/10.1162/tacl_a_00261} {{SECTOR}: A neural model for coherent topic segmentation and classification}.
\newblock \emph{Transactions of the Association for Computational Linguistics}, 7:169--184.

\bibitem[{Blei et~al.(2003)Blei, Ng, and Jordan}]{blei2003latent}
David~M Blei, Andrew~Y Ng, and Michael~I Jordan. 2003.
\newblock Latent dirichlet allocation.
\newblock \emph{Journal of machine Learning research}, 3(Jan):993--1022.

\bibitem[{Bubeck et~al.(2023)Bubeck, Chandrasekaran, Eldan, Gehrke, Horvitz, Kamar, Lee, Lee, Li, Lundberg, Nori, Palangi, Ribeiro, and Zhang}]{bubeck2023sparksartificialgeneralintelligence}
Sébastien Bubeck, Varun Chandrasekaran, Ronen Eldan, Johannes Gehrke, Eric Horvitz, Ece Kamar, Peter Lee, Yin~Tat Lee, Yuanzhi Li, Scott Lundberg, Harsha Nori, Hamid Palangi, Marco~Tulio Ribeiro, and Yi~Zhang. 2023.
\newblock \href {https://arxiv.org/abs/2303.12712} {Sparks of artificial general intelligence: Early experiments with gpt-4}.
\newblock \emph{Preprint}, arXiv:2303.12712.

\bibitem[{Deng et~al.(2024)Deng, Zhao, Tang, Gerstein, and Cohan}]{deng-etal-2024-investigating}
Chunyuan Deng, Yilun Zhao, Xiangru Tang, Mark Gerstein, and Arman Cohan. 2024.
\newblock \href {https://doi.org/10.18653/v1/2024.naacl-long.482} {Investigating data contamination in modern benchmarks for large language models}.
\newblock In \emph{Proceedings of the 2024 Conference of the North American Chapter of the Association for Computational Linguistics: Human Language Technologies (Volume 1: Long Papers)}, pages 8706--8719, Mexico City, Mexico. Association for Computational Linguistics.

\bibitem[{Dieng et~al.(2020)Dieng, Ruiz, and Blei}]{ETM}
Adji~B. Dieng, Francisco J.~R. Ruiz, and David~M. Blei. 2020.
\newblock \href {https://doi.org/10.1162/tacl_a_00325} {{Topic Modeling in Embedding Spaces}}.
\newblock \emph{Transactions of the Association for Computational Linguistics}, 8:439--453.

\bibitem[{Dubey et~al.(2024)Dubey, Jauhri, Pandey, Kadian, Al-Dahle, Letman, Mathur, Schelten, Yang, Fan et~al.}]{dubey2024llama}
Abhimanyu Dubey, Abhinav Jauhri, Abhinav Pandey, Abhishek Kadian, Ahmad Al-Dahle, Aiesha Letman, Akhil Mathur, Alan Schelten, Amy Yang, Angela Fan, et~al. 2024.
\newblock The llama 3 herd of models.
\newblock \emph{arXiv preprint arXiv:2407.21783}.

\bibitem[{Fan et~al.(2018)Fan, Lewis, and Dauphin}]{fan-etal-2018-hierarchical}
Angela Fan, Mike Lewis, and Yann Dauphin. 2018.
\newblock \href {https://doi.org/10.18653/v1/P18-1082} {Hierarchical neural story generation}.
\newblock In \emph{Proceedings of the 56th Annual Meeting of the Association for Computational Linguistics (Volume 1: Long Papers)}, pages 889--898, Melbourne, Australia. Association for Computational Linguistics.

\bibitem[{Gillings and Hardie(2022)}]{Gillings2022}
Mathew Gillings and Andrew Hardie. 2022.
\newblock \href {https://doi.org/10.1093/llc/fqac075} {{The interpretation of topic models for scholarly analysis: An evaluation and critique of current practice}}.
\newblock \emph{Digital Scholarship in the Humanities}, 38(2):530--543.

\bibitem[{Glava{\v{s}} et~al.(2016)Glava{\v{s}}, Nanni, and Ponzetto}]{glavas-etal-2016-unsupervised}
Goran Glava{\v{s}}, Federico Nanni, and Simone~Paolo Ponzetto. 2016.
\newblock \href {https://doi.org/10.18653/v1/S16-2016} {Unsupervised text segmentation using semantic relatedness graphs}.
\newblock In \emph{Proceedings of the Fifth Joint Conference on Lexical and Computational Semantics}, pages 125--130, Berlin, Germany. Association for Computational Linguistics.

\bibitem[{Guan et~al.(2025)Guan, Zhang, Liu, Shang, Sun, Zhu, Yang, and Yang}]{guan2025rstarmathsmallllmsmaster}
Xinyu Guan, Li~Lyna Zhang, Yifei Liu, Ning Shang, Youran Sun, Yi~Zhu, Fan Yang, and Mao Yang. 2025.
\newblock \href {https://arxiv.org/abs/2501.04519} {rstar-math: Small llms can master math reasoning with self-evolved deep thinking}.
\newblock \emph{Preprint}, arXiv:2501.04519.

\bibitem[{Hearst(1997)}]{hearst1997text}
Marti~A. Hearst. 1997.
\newblock \href {https://aclanthology.org/J97-1003} {Text tiling: Segmenting text into multi-paragraph subtopic passages}.
\newblock \emph{Computational Linguistics}, 23(1):33--64.

\bibitem[{Hendrycks et~al.(2021)Hendrycks, Burns, Kadavath, Arora, Basart, Tang, Song, and Steinhardt}]{hendrycks2021math}
Dan Hendrycks, Collin Burns, Saurav Kadavath, Akul Arora, Steven Basart, Eric Tang, Dawn Song, and Jacob Steinhardt. 2021.
\newblock \href {https://arxiv.org/abs/2103.03874} {Measuring mathematical problem solving with the {MATH} dataset}.
\newblock \emph{CoRR}, abs/2103.03874.

\bibitem[{Hofmann(1999)}]{hofmann1999lsi}
T~Hofmann. 1999.
\newblock Probabilistic latent semantic indexing.
\newblock In \emph{Proceedings of the 22nd annual international ACM SIGIR conference on Research and development in information retrieval}.

\bibitem[{Hofmann(2001)}]{hofmann2001unsupervised}
Thomas Hofmann. 2001.
\newblock Unsupervised learning by probabilistic latent semantic analysis.
\newblock \emph{Machine learning}, 42:177--196.

\bibitem[{Hofmann et~al.(1999)}]{hofmann1999lsa}
Thomas Hofmann et~al. 1999.
\newblock Probabilistic latent semantic analysis.
\newblock In \emph{UAI}, volume~99, pages 289--296.

\bibitem[{Katz et~al.(2024)Katz, Bommarito, Gao, and Arredondo}]{katz2024gpt}
Daniel~Martin Katz, Michael~James Bommarito, Shang Gao, and Pablo Arredondo. 2024.
\newblock Gpt-4 passes the bar exam.
\newblock \emph{Philosophical Transactions of the Royal Society A}, 382(2270):20230254.

\bibitem[{Koshorek et~al.(2018)Koshorek, Cohen, Mor, Rotman, and Berant}]{koshorek-etal-2018-text}
Omri Koshorek, Adir Cohen, Noam Mor, Michael Rotman, and Jonathan Berant. 2018.
\newblock \href {https://doi.org/10.18653/v1/N18-2075} {Text segmentation as a supervised learning task}.
\newblock In \emph{Proceedings of the 2018 Conference of the North {A}merican Chapter of the Association for Computational Linguistics: Human Language Technologies, Volume 2 (Short Papers)}, pages 469--473, New Orleans, Louisiana. Association for Computational Linguistics.

\bibitem[{Lee et~al.(2017)Lee, Smith, Seppi, Elmqvist, Boyd-Graber, and Findlater}]{LEE201728}
Tak~Yeon Lee, Alison Smith, Kevin Seppi, Niklas Elmqvist, Jordan Boyd-Graber, and Leah Findlater. 2017.
\newblock \href {https://doi.org/10.1016/j.ijhcs.2017.03.007} {The human touch: How non-expert users perceive, interpret, and fix topic models}.
\newblock \emph{International Journal of Human-Computer Studies}, 105:28--42.

\bibitem[{Li et~al.(2023)Li, Zhang, and Zhou}]{LiZZ2023}
Dai Li, Bolun Zhang, and Yimang Zhou. 2023.
\newblock \href {https://doi.org/10.31235/osf.io/23x4m} {Can large language models (llm) label topics from a topic model?}

\bibitem[{Liu et~al.(2023)Liu, Ning, Teng, Liu, Zhou, and Zhang}]{liu2023evaluating}
Hanmeng Liu, Ruoxi Ning, Zhiyang Teng, Jian Liu, Qiji Zhou, and Yue Zhang. 2023.
\newblock Evaluating the logical reasoning ability of chatgpt and gpt-4.
\newblock \emph{arXiv preprint arXiv:2304.03439}.

\bibitem[{Miao et~al.(2016)Miao, Yu, and Blunsom}]{NVDM}
Yishu Miao, Lei Yu, and Phil Blunsom. 2016.
\newblock \href {https://proceedings.mlr.press/v48/miao16.html} {Neural variational inference for text processing}.
\newblock In \emph{Proceedings of The 33rd International Conference on Machine Learning}, volume~48 of \emph{Proceedings of Machine Learning Research}, pages 1727--1736, New York, New York, USA. PMLR.

\bibitem[{Mu et~al.(2024)Mu, Dong, Bontcheva, and Song}]{mu-etal-2024-large}
Yida Mu, Chun Dong, Kalina Bontcheva, and Xingyi Song. 2024.
\newblock \href {https://aclanthology.org/2024.lrec-main.887} {Large language models offer an alternative to the traditional approach of topic modelling}.
\newblock In \emph{Proceedings of the 2024 Joint International Conference on Computational Linguistics, Language Resources and Evaluation (LREC-COLING 2024)}, pages 10160--10171, Torino, Italia. ELRA and ICCL.

\bibitem[{OpenAI(2023)}]{openai2023gpt4}
OpenAI. 2023.
\newblock \href {https://arxiv.org/abs/2303.08774} {Gpt-4 technical report}.
\newblock \emph{Preprint}, arXiv:2303.08774.

\bibitem[{OpenAI et~al.(2024)OpenAI, Achiam, Adler, Agarwal, Ahmad, Akkaya, Aleman, Almeida, Altenschmidt, Altman, Anadkat, Avila, Babuschkin, Balaji, Balcom, Baltescu, Bao, Bavarian, Belgum, Bello, Berdine, Bernadett-Shapiro, Berner, Bogdonoff, Boiko, Boyd, Brakman, Brockman, Brooks, Brundage, Button, Cai, Campbell, Cann, Carey, Carlson, Carmichael, Chan, Chang, Chantzis, Chen, Chen, Chen, Chen, Chen, Chess, Cho, Chu, Chung, Cummings, Currier, Dai, Decareaux, Degry, Deutsch, Deville, Dhar, Dohan, Dowling, Dunning, Ecoffet, Eleti, Eloundou, Farhi, Fedus, Felix, Fishman, Forte, Fulford, Gao, Georges, Gibson, Goel, Gogineni, Goh, Gontijo-Lopes, Gordon, Grafstein, Gray, Greene, Gross, Gu, Guo, Hallacy, Han, Harris, He, Heaton, Heidecke, Hesse, Hickey, Hickey, Hoeschele, Houghton, Hsu, Hu, Hu, Huizinga, Jain, Jain, Jang, Jiang, Jiang, Jin, Jin, Jomoto, Jonn, Jun, Kaftan, Łukasz Kaiser, Kamali, Kanitscheider, Keskar, Khan, Kilpatrick, Kim, Kim, Kim, Kirchner, Kiros, Knight, Kokotajlo, Łukasz Kondraciuk,
  Kondrich, Konstantinidis, Kosic, Krueger, Kuo, Lampe, Lan, Lee, Leike, Leung, Levy, Li, Lim, Lin, Lin, Litwin, Lopez, Lowe, Lue, Makanju, Malfacini, Manning, Markov, Markovski, Martin, Mayer, Mayne, McGrew, McKinney, McLeavey, McMillan, McNeil, Medina, Mehta, Menick, Metz, Mishchenko, Mishkin, Monaco, Morikawa, Mossing, Mu, Murati, Murk, Mély, Nair, Nakano, Nayak, Neelakantan, Ngo, Noh, Ouyang, O'Keefe, Pachocki, Paino, Palermo, Pantuliano, Parascandolo, Parish, Parparita, Passos, Pavlov, Peng, Perelman, de~Avila Belbute~Peres, Petrov, de~Oliveira~Pinto, Michael, Pokorny, Pokrass, Pong, Powell, Power, Power, Proehl, Puri, Radford, Rae, Ramesh, Raymond, Real, Rimbach, Ross, Rotsted, Roussez, Ryder, Saltarelli, Sanders, Santurkar, Sastry, Schmidt, Schnurr, Schulman, Selsam, Sheppard, Sherbakov, Shieh, Shoker, Shyam, Sidor, Sigler, Simens, Sitkin, Slama, Sohl, Sokolowsky, Song, Staudacher, Such, Summers, Sutskever, Tang, Tezak, Thompson, Tillet, Tootoonchian, Tseng, Tuggle, Turley, Tworek, Uribe, Vallone,
  Vijayvergiya, Voss, Wainwright, Wang, Wang, Wang, Ward, Wei, Weinmann, Welihinda, Welinder, Weng, Weng, Wiethoff, Willner, Winter, Wolrich, Wong, Workman, Wu, Wu, Wu, Xiao, Xu, Yoo, Yu, Yuan, Zaremba, Zellers, Zhang, Zhang, Zhao, Zheng, Zhuang, Zhuk, and Zoph}]{openai2024gpt4technicalreport}
OpenAI, Josh Achiam, Steven Adler, Sandhini Agarwal, Lama Ahmad, Ilge Akkaya, Florencia~Leoni Aleman, Diogo Almeida, Janko Altenschmidt, Sam Altman, Shyamal Anadkat, Red Avila, Igor Babuschkin, Suchir Balaji, Valerie Balcom, Paul Baltescu, Haiming Bao, Mohammad Bavarian, Jeff Belgum, Irwan Bello, Jake Berdine, Gabriel Bernadett-Shapiro, Christopher Berner, Lenny Bogdonoff, Oleg Boiko, Madelaine Boyd, Anna-Luisa Brakman, Greg Brockman, Tim Brooks, Miles Brundage, Kevin Button, Trevor Cai, Rosie Campbell, Andrew Cann, Brittany Carey, Chelsea Carlson, Rory Carmichael, Brooke Chan, Che Chang, Fotis Chantzis, Derek Chen, Sully Chen, Ruby Chen, Jason Chen, Mark Chen, Ben Chess, Chester Cho, Casey Chu, Hyung~Won Chung, Dave Cummings, Jeremiah Currier, Yunxing Dai, Cory Decareaux, Thomas Degry, Noah Deutsch, Damien Deville, Arka Dhar, David Dohan, Steve Dowling, Sheila Dunning, Adrien Ecoffet, Atty Eleti, Tyna Eloundou, David Farhi, Liam Fedus, Niko Felix, Simón~Posada Fishman, Juston Forte, Isabella Fulford, Leo
  Gao, Elie Georges, Christian Gibson, Vik Goel, Tarun Gogineni, Gabriel Goh, Rapha Gontijo-Lopes, Jonathan Gordon, Morgan Grafstein, Scott Gray, Ryan Greene, Joshua Gross, Shixiang~Shane Gu, Yufei Guo, Chris Hallacy, Jesse Han, Jeff Harris, Yuchen He, Mike Heaton, Johannes Heidecke, Chris Hesse, Alan Hickey, Wade Hickey, Peter Hoeschele, Brandon Houghton, Kenny Hsu, Shengli Hu, Xin Hu, Joost Huizinga, Shantanu Jain, Shawn Jain, Joanne Jang, Angela Jiang, Roger Jiang, Haozhun Jin, Denny Jin, Shino Jomoto, Billie Jonn, Heewoo Jun, Tomer Kaftan, Łukasz Kaiser, Ali Kamali, Ingmar Kanitscheider, Nitish~Shirish Keskar, Tabarak Khan, Logan Kilpatrick, Jong~Wook Kim, Christina Kim, Yongjik Kim, Jan~Hendrik Kirchner, Jamie Kiros, Matt Knight, Daniel Kokotajlo, Łukasz Kondraciuk, Andrew Kondrich, Aris Konstantinidis, Kyle Kosic, Gretchen Krueger, Vishal Kuo, Michael Lampe, Ikai Lan, Teddy Lee, Jan Leike, Jade Leung, Daniel Levy, Chak~Ming Li, Rachel Lim, Molly Lin, Stephanie Lin, Mateusz Litwin, Theresa Lopez, Ryan
  Lowe, Patricia Lue, Anna Makanju, Kim Malfacini, Sam Manning, Todor Markov, Yaniv Markovski, Bianca Martin, Katie Mayer, Andrew Mayne, Bob McGrew, Scott~Mayer McKinney, Christine McLeavey, Paul McMillan, Jake McNeil, David Medina, Aalok Mehta, Jacob Menick, Luke Metz, Andrey Mishchenko, Pamela Mishkin, Vinnie Monaco, Evan Morikawa, Daniel Mossing, Tong Mu, Mira Murati, Oleg Murk, David Mély, Ashvin Nair, Reiichiro Nakano, Rajeev Nayak, Arvind Neelakantan, Richard Ngo, Hyeonwoo Noh, Long Ouyang, Cullen O'Keefe, Jakub Pachocki, Alex Paino, Joe Palermo, Ashley Pantuliano, Giambattista Parascandolo, Joel Parish, Emy Parparita, Alex Passos, Mikhail Pavlov, Andrew Peng, Adam Perelman, Filipe de~Avila Belbute~Peres, Michael Petrov, Henrique~Ponde de~Oliveira~Pinto, Michael, Pokorny, Michelle Pokrass, Vitchyr~H. Pong, Tolly Powell, Alethea Power, Boris Power, Elizabeth Proehl, Raul Puri, Alec Radford, Jack Rae, Aditya Ramesh, Cameron Raymond, Francis Real, Kendra Rimbach, Carl Ross, Bob Rotsted, Henri Roussez,
  Nick Ryder, Mario Saltarelli, Ted Sanders, Shibani Santurkar, Girish Sastry, Heather Schmidt, David Schnurr, John Schulman, Daniel Selsam, Kyla Sheppard, Toki Sherbakov, Jessica Shieh, Sarah Shoker, Pranav Shyam, Szymon Sidor, Eric Sigler, Maddie Simens, Jordan Sitkin, Katarina Slama, Ian Sohl, Benjamin Sokolowsky, Yang Song, Natalie Staudacher, Felipe~Petroski Such, Natalie Summers, Ilya Sutskever, Jie Tang, Nikolas Tezak, Madeleine~B. Thompson, Phil Tillet, Amin Tootoonchian, Elizabeth Tseng, Preston Tuggle, Nick Turley, Jerry Tworek, Juan Felipe~Cerón Uribe, Andrea Vallone, Arun Vijayvergiya, Chelsea Voss, Carroll Wainwright, Justin~Jay Wang, Alvin Wang, Ben Wang, Jonathan Ward, Jason Wei, CJ~Weinmann, Akila Welihinda, Peter Welinder, Jiayi Weng, Lilian Weng, Matt Wiethoff, Dave Willner, Clemens Winter, Samuel Wolrich, Hannah Wong, Lauren Workman, Sherwin Wu, Jeff Wu, Michael Wu, Kai Xiao, Tao Xu, Sarah Yoo, Kevin Yu, Qiming Yuan, Wojciech Zaremba, Rowan Zellers, Chong Zhang, Marvin Zhang, Shengjia
  Zhao, Tianhao Zheng, Juntang Zhuang, William Zhuk, and Barret Zoph. 2024.
\newblock \href {https://arxiv.org/abs/2303.08774} {Gpt-4 technical report}.
\newblock \emph{Preprint}, arXiv:2303.08774.

\bibitem[{Pham et~al.(2024)Pham, Hoyle, Sun, Resnik, and Iyyer}]{pham-etal-2024-topicgpt}
Chau Pham, Alexander Hoyle, Simeng Sun, Philip Resnik, and Mohit Iyyer. 2024.
\newblock \href {https://doi.org/10.18653/v1/2024.naacl-long.164} {{T}opic{GPT}: A prompt-based topic modeling framework}.
\newblock In \emph{Proceedings of the 2024 Conference of the North American Chapter of the Association for Computational Linguistics: Human Language Technologies (Volume 1: Long Papers)}, pages 2956--2984, Mexico City, Mexico. Association for Computational Linguistics.

\bibitem[{Qwen et~al.(2025)Qwen, :, Yang, Yang, Zhang, Hui, Zheng, Yu, Li, Liu, Huang, Wei, Lin, Yang, Tu, Zhang, Yang, Yang, Zhou, Lin, Dang, Lu, Bao, Yang, Yu, Li, Xue, Zhang, Zhu, Men, Lin, Li, Tang, Xia, Ren, Ren, Fan, Su, Zhang, Wan, Liu, Cui, Zhang, and Qiu}]{qwen2025qwen25technicalreport}
Qwen, :, An~Yang, Baosong Yang, Beichen Zhang, Binyuan Hui, Bo~Zheng, Bowen Yu, Chengyuan Li, Dayiheng Liu, Fei Huang, Haoran Wei, Huan Lin, Jian Yang, Jianhong Tu, Jianwei Zhang, Jianxin Yang, Jiaxi Yang, Jingren Zhou, Junyang Lin, Kai Dang, Keming Lu, Keqin Bao, Kexin Yang, Le~Yu, Mei Li, Mingfeng Xue, Pei Zhang, Qin Zhu, Rui Men, Runji Lin, Tianhao Li, Tianyi Tang, Tingyu Xia, Xingzhang Ren, Xuancheng Ren, Yang Fan, Yang Su, Yichang Zhang, Yu~Wan, Yuqiong Liu, Zeyu Cui, Zhenru Zhang, and Zihan Qiu. 2025.
\newblock \href {https://arxiv.org/abs/2412.15115} {Qwen2.5 technical report}.
\newblock \emph{Preprint}, arXiv:2412.15115.

\bibitem[{Riedl and Biemann(2012)}]{riedl-biemann-2012-topictiling}
Martin Riedl and Chris Biemann. 2012.
\newblock \href {https://aclanthology.org/W12-3307} {{T}opic{T}iling: A text segmentation algorithm based on {LDA}}.
\newblock In \emph{Proceedings of {ACL} 2012 Student Research Workshop}, pages 37--42, Jeju Island, Korea. Association for Computational Linguistics.

\bibitem[{Srivastava and Sutton(2017)}]{srivastava2017autoencodingvariationalinferencetopic}
Akash Srivastava and Charles Sutton. 2017.
\newblock \href {https://arxiv.org/abs/1703.01488} {Autoencoding variational inference for topic models}.
\newblock \emph{Preprint}, arXiv:1703.01488.

\bibitem[{Suzgun et~al.(2022)Suzgun, Scales, Sch{\"a}rli, Gehrmann, Tay, Chung, Chowdhery, Le, Chi, Zhou et~al.}]{suzgun2022challenging}
Mirac Suzgun, Nathan Scales, Nathanael Sch{\"a}rli, Sebastian Gehrmann, Yi~Tay, Hyung~Won Chung, Aakanksha Chowdhery, Quoc~V Le, Ed~H Chi, Denny Zhou, et~al. 2022.
\newblock Challenging big-bench tasks and whether chain-of-thought can solve them.
\newblock \emph{arXiv preprint arXiv:2210.09261}.

\bibitem[{Wang et~al.(2023)Wang, Prakash, Hoang, Hee, Naseem, and Lee}]{Wang2023PromptTopic}
Han Wang, Nirmalendu Prakash, Nguyen~Khoi Hoang, Ming~Shan Hee, Usman Naseem, and Roy Ka-Wei Lee. 2023.
\newblock \href {https://doi.org/10.1109/BigData59044.2023.10386113} {Prompting large language models for topic modeling}.
\newblock In \emph{2023 IEEE International Conference on Big Data (BigData)}, pages 1236--1241.

\bibitem[{Wei et~al.(2022)Wei, Wang, Schuurmans, Bosma, Xia, Chi, Le, Zhou et~al.}]{wei2022chain}
Jason Wei, Xuezhi Wang, Dale Schuurmans, Maarten Bosma, Fei Xia, Ed~Chi, Quoc~V Le, Denny Zhou, et~al. 2022.
\newblock Chain-of-thought prompting elicits reasoning in large language models.
\newblock \emph{Advances in Neural Information Processing Systems}, 35:24824--24837.

\bibitem[{Wu et~al.(2023)Wu, Jia, Zhang, Li, Zhu, Wang, Lee, Peng, Wu, and Wang}]{wu2023empirical}
Yiran Wu, Feiran Jia, Shaokun Zhang, Hangyu Li, Erkang Zhu, Yue Wang, Yin~Tat Lee, Richard Peng, Qingyun Wu, and Chi Wang. 2023.
\newblock An empirical study on challenging math problem solving with gpt-4.
\newblock \emph{arXiv preprint arXiv:2306.01337}.

\bibitem[{Xu et~al.(2024)Xu, Banburski, and Jojic}]{Xu2024Reprompting}
Weijia Xu, Andrzej Banburski, and Nebojsa Jojic. 2024.
\newblock \href {https://proceedings.mlr.press/v235/xu24b.html} {Reprompting: Automated chain-of-thought prompt inference through {G}ibbs sampling}.
\newblock In \emph{Proceedings of the 41st International Conference on Machine Learning}, volume 235 of \emph{Proceedings of Machine Learning Research}, pages 54852--54865. PMLR.

\bibitem[{Zhang et~al.(2019)Zhang, Guo, Fan, Lan, and Cheng}]{zhang2019outline}
Ruqing Zhang, Jiafeng Guo, Yixing Fan, Yanyan Lan, and Xueqi Cheng. 2019.
\newblock Outline generation: Understanding the inherent content structure of documents.
\newblock In \emph{Proceedings of the 42nd International ACM SIGIR Conference on Research and Development in Information Retrieval}, pages 745--754.

\bibitem[{Zhou et~al.(2023)Zhou, Wang, Lu, Shi, Luo, Qin, Lu, Jia, Song, Zhan, and Li}]{zhou2023solvingchallengingmathword}
Aojun Zhou, Ke~Wang, Zimu Lu, Weikang Shi, Sichun Luo, Zipeng Qin, Shaoqing Lu, Anya Jia, Linqi Song, Mingjie Zhan, and Hongsheng Li. 2023.
\newblock \href {https://arxiv.org/abs/2308.07921} {Solving challenging math word problems using gpt-4 code interpreter with code-based self-verification}.
\newblock \emph{Preprint}, arXiv:2308.07921.

\end{thebibliography}

\appendix
\onecolumn
\section{Appendix}

\subsection{Datasets}
We evaluate \codename against various baselines on story writing, math problem solving and multi-step reasoning benchmarks. We use WritingPrompts~\citep{fan-etal-2018-hierarchical}, a story writing dataset that contains 300K human-written stories paired with writing prompts from an online forum. We randomly sample~100 stories from the training set for clustering and tagging. We set the number of tags to~100 for all tagging approaches.
For math problem solving, we use MATH~\citep{hendrycks2021math}, a popular math benchmark that contains high school math competition problems on seven subjects including Prealgebra, Algebra, Number Theory, Counting and Probability, Geometry, Intermediate Algebra
and Precalculus. We learn~100 tags on~1K randomly sampled problem solutions from the training set. We also experiment on the Big-Bench Hard~(BBH) benchmark~\citep{suzgun2022challenging}. The original benchmark includes~23 challenging multi-step reasoning tasks, but each task only includes three step-by-step solution examples. Instead, we take the~12 tasks used in~\citet{Xu2024Reprompting} and learn the tags on the problem solutions~(produced by their automatic prompt inference algorithm) for the~179 training problems. We set the number of tags to~50 for BBH.\footnote{All datasets used in the work are under MIT license. Our use of the datasets is consistent with their intended use.}

\subsection{Large Language Model Setup}
For clustering and tagging, we use GPT-4~\citep{openai2024gpt4technicalreport} and Qwen-2.5-7B~(a much smaller LLM introduced in~\citet{qwen2025qwen25technicalreport}). For GPT-4, we set~$top\_p = 0.5$, sampling temperature~$\tau = 1.0$, zero frequency and presence penalty. For Qwen-2.5-7B, we set~$top\_p = 0.5$, sampling temperature~$\tau = 0.1$, zero frequency and presence penalty.

We also use GPT-4 with~$top\_p = 0.5$ to estimate the reconstruction log-likelihood. We set the temperature~$\tau = 1.0$ when sampling alternative segments and~$\tau = 0$ when choosing the best continuation.

To measure Hits@K Accuracy, we use ChatGPT~(gpt-3.5-turbo; \citet{openai2023gpt4}) instead of GPT-4.
We set~$top\_p = 0.5$ and temperature~$\tau = 1.0$ when sampling solutions from ChatGPT.

\subsection{Ablation Study}
\ref{tab:ablation_results} shows the ablation study results on the number of tags.

\begin{table}[!ht]
    \centering
    \begin{tabular}{l|ccc}
        ~ & 20 Tags & 50 Tags & 100 Tags  \\ \midrule
        Date & 98.0 & 98.8 & 99.2  \\ 
        Formal & 63.2 & 93.2 & 80.8  \\ 
        Geometric & 86.4 & 87.6 & 86.0  \\ 
        Logical & 98.9 & 99.5 & 99.1  \\ 
        Movie & 93.6 & 95.2 & 94.8  \\ 
        ObjCount & 99.6 & 95.2 & 99.6  \\ 
        Penguins & 99.3 & 99.3 & 99.3  \\ 
        ReasonColored & 100.0 & 100.0 & 100.0  \\ 
        RuinNames & 90.8 & 93.6 & 95.6  \\ 
        TranslationError & 72.8 & 75.2 & 72.4  \\ 
        Temporal & 98.8 & 100.0 & 99.2  \\ 
        WordSort & 57.6 & 56.0 & 61.6  \\\midrule
        {\bf Average} & 88.3 & {\bf 91.1} & 90.6 \\
        \midrule
    \end{tabular}
\caption{Ablation Study: Hits@K Accuracy on BBH tasks using varying number of \codename tags.}
\label{tab:ablation_results}
\end{table}

\end{document}